\pdfoutput=1
\documentclass[10pt,twocolumn,letterpaper]{article}
\usepackage{titling}
\usepackage{iccv}
\usepackage{times}
\usepackage{epsfig}
\usepackage{graphicx}
\usepackage{amsmath}
\usepackage{amssymb}

\usepackage{booktabs}
\usepackage{subfig}
\usepackage{bibunits}

\usepackage[pagebackref=true,breaklinks=true,letterpaper=true,colorlinks,bookmarks=false]{hyperref}

\iccvfinalcopy 

\def\iccvPaperID{2081} 

\graphicspath{{figures/}}
\DeclareMathAlphabet{\pazocal}{OMS}{zplm}{m}{n}
\newcommand{\neigh}[1]{\mathcal{N}_i}
\newcommand{\x}{\textbf{x}}
\renewcommand{\xi}{\textbf{x}_i}
\renewcommand{\t}{\textbf{t}}
\newcommand{\m}{\textbf{m}}
\newcommand{\mi}{\textbf{m}_i}
\newcommand{\mij}{\textbf{m}_{i,j}}
\newcommand{\miik}{\textbf{m}_{i,i-k}}
\newcommand{\mii}{\textbf{m}_{i,i}}
\newcommand{\lij}{\ell_{i,j}}
\newcommand{\xj}{\textbf{x}_j}
\newcommand{\e}{\textbf{e}}
\newcommand{\ei}{\textbf{e}_i}
\newcommand{\ej}{\textbf{e}_j}

\newcommand{\yi}{\textbf{y}_i}

\newcommand{\li}{\textbf{l}_i}
\newcommand{\p}{\textbf{p}}
\renewcommand{\pi}{\textbf{p}_i}
\newcommand{\pj}{\textbf{p}_j}
\newcommand{\lj}{\textbf{l}_j}

\newcommand{\I}{\textbf{I}}
\newcommand{\R}{\mathbb{R}}

\newcommand{\refeq}[1]{Eq.~\ref{#1}}

\newcommand{\refsec}[1]{Sec.~\ref{#1}}
\newcommand{\reffig}[1]{Figure~\ref{#1}}

\begin{document}

\title{Segmentation-Aware Convolutional Networks Using Local Attention Masks}

\author{Adam W. Harley\\
Carnegie Mellon University\\
{\tt\small aharley@cmu.edu}
\and
Konstantinos G. Derpanis\\
Ryerson University\\
{\tt\small kosta@ryerson.ca}
\and
Iasonas Kokkinos\\
Facebook AI Research\\
{\tt\small iasonask@fb.com}
}

\maketitle


\begin{abstract}
We introduce an approach to integrate segmentation information within a convolutional neural network (CNN). This counter-acts the tendency of CNNs to smooth information across regions and increases their spatial precision. To obtain segmentation information, we set up a CNN to provide an embedding space where region co-membership can be estimated based on Euclidean distance. We use these embeddings to compute a local attention mask relative to every neuron position. We incorporate such masks in CNNs and replace the convolution operation with a ``segmentation-aware'' variant that allows a neuron to selectively attend to inputs coming from its own region. We call the resulting network a segmentation-aware CNN because it adapts its filters at each image point according to local segmentation cues. We demonstrate the merit of our method on two widely different dense prediction tasks, that involve classification (semantic segmentation) and regression (optical flow). Our results show that in semantic segmentation we can match the performance of DenseCRFs while being faster and simpler, and in optical flow we obtain clearly sharper responses than networks that do not use local attention masks. In both cases, segmentation-aware convolution yields systematic improvements over strong baselines. Source code for this work is available online at \url{http://cs.cmu.edu/~aharley/segaware}.
  
\end{abstract}

\section{Introduction}

Convolutional neural networks (CNNs) have recently made rapid progress in pixel-wise prediction tasks, including depth prediction \cite{eigen2014depth}, optical flow estimation \cite{flownet}, and semantic segmentation \cite{sermanet-iclr-14,chen_deeplab,long_shelhamer_fcn}. This progress has been built on the remarkable success of CNNs in image classification tasks \cite{kriz,Simonyan14c} -- indeed, most dense prediction models are based closely on architectures that were successful in object recognition. While this strategy facilitates transfer learning, it also brings design elements that are incompatible with dense prediction. 

\begin{figure}[t]
\begin{center}
  \includegraphics[width=1.0\linewidth]{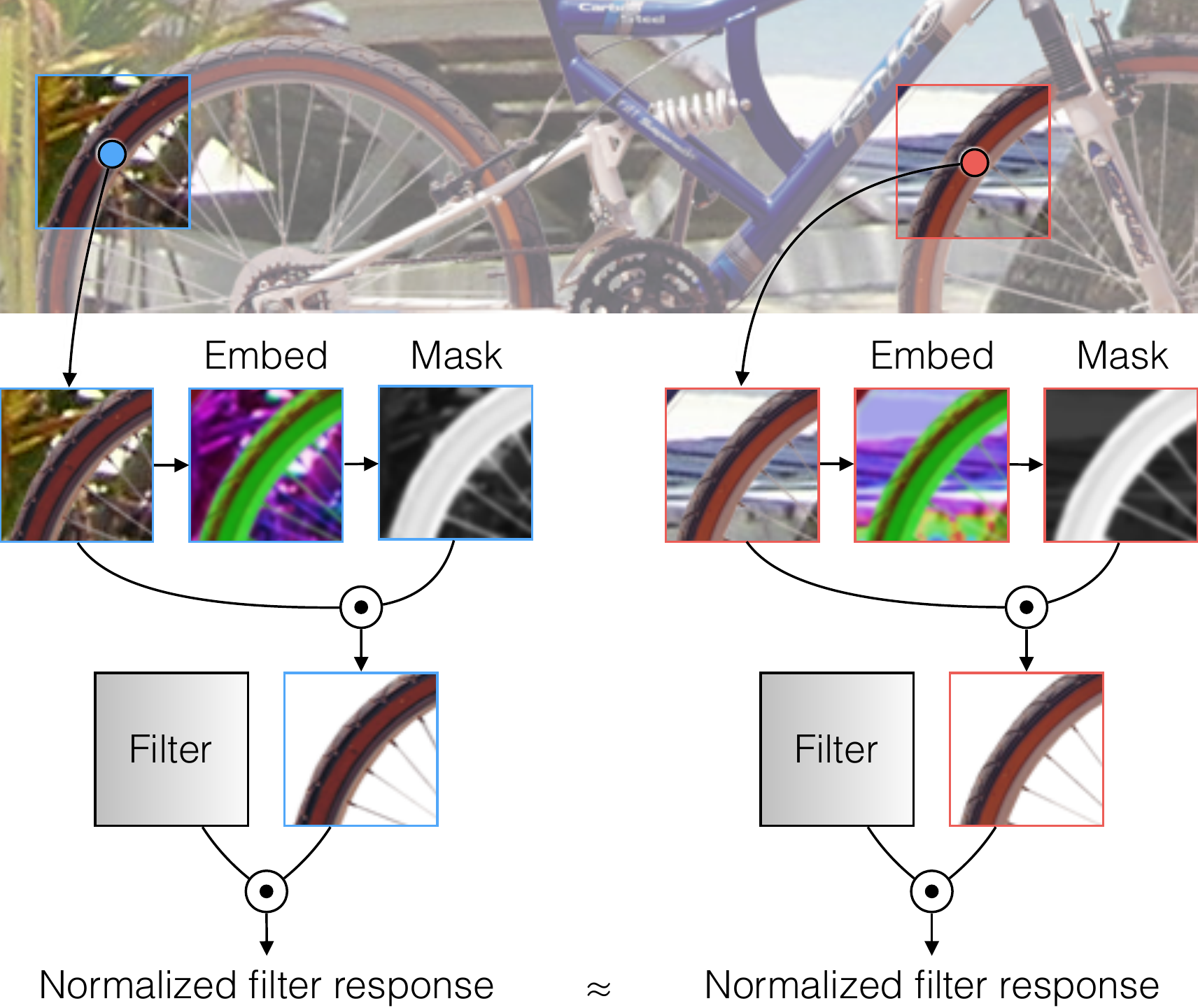}
\end{center}
\vspace{-12pt}
\caption{Segmentation-aware convolution filters are invariant to backgrounds. We achieve this in three steps: (i) compute segmentation cues for each pixel (\ie, ``embeddings''), (ii) create a foreground mask for each patch, and (iii) combine the masks with convolution, so that the filters only process the local foreground in each image patch.}
\label{fig:segaware}
\vspace{-12pt}
\end{figure}

By design CNNs typically produce feature maps and predictions that are smooth and low-resolution, resulting from the repeated pooling and subsampling stages in the network architecture, respectively. These stages play an important role in the hierarchical consolidation of features, and widen the higher layer effective receptive fields.  The low-resolution issue has received substantial attention: for instance  methods have been proposed for replacing the subsampling layers with resolution-preserving alternatives such as atrous convolution \cite{chen_deeplab, fisher, pohlen2016full}, or restoring the lost resolution via upsampling stages \cite{noh2015learning,long_shelhamer_fcn}.
However, the issue of smoothness has remained relatively unexplored. Smooth neuron outputs result from the spatial  pooling (\ie, abstraction) of information across different regions.
This can be useful in high-level tasks, but can
degrade accuracy on per-pixel prediction tasks where rapid changes in activation may be required, \eg, around region boundaries or motion discontinuities.



To address the issue of smoothness, we propose \textit{segmentation-aware} convolutional networks, which operate as illustrated in \reffig{fig:segaware}.
These networks adjust their behavior on a \textit{per-pixel} basis according to segmentation cues, so that the filters can selectively ``attend'' to information coming from the region containing the neuron, and treat it differently from background signals. 
To achieve this, we complement each image patch with a local foreground-background segmentation mask that acts like a gating mechanism for the information feeding into the neuron. This avoids feature blurring, by reducing the extent to which foreground and contextual information is mixed, and allows neuron  
 activation levels to change rapidly, by dynamically adapting the neuron's behavior to the image content. 
This goes beyond sharpening the network outputs post-hoc, as is currently common practice; it fixes the blurring problem ``before the damage is done'', since it can be integrated at both early and later stages of a CNN.

The general idea of combining filtering with  segmentation to enhance sharpness dates back to nonlinear image processing \cite{PeronaM90,tomasi1998bilateral} and segmentation-aware feature extraction \cite{trulls2013dense,trulls2014segmentation}. Apart from showing that this technique successfully carries over to CNNs, another contribution of our work consists in using the network itself to obtain segmentation information, rather than relying on hand-crafted pipelines. In particular, as in an earlier version of this work \cite{harley2016iclr}, we use a constrastive side loss to train the ``segmentation embedding''  branch of our network, so that we can then  construct segmentation masks using embedding distances. 






There are three steps to creating segmentation-aware convolutional nets, described in Sections 3.1-3.4: (i) learn segmentation cues, (ii) use the cues to create local foreground masks, and (iii) use the masks together with convolution, to create foreground-focused convolution. 
Our approach realizes each of these steps in a unified manner that is at once general (\ie, applicable to both discrete and continuous prediction tasks), differentiable (\ie, end-to-end trainable as a neural
network), and fast (\ie, implemented  as GPU-optimized variants of convolution).

Experiments show that minimally modifying existing CNN architectures to use segmentation-aware convolution yields substantial gains in two widely different task settings: dense discrete labelling (\ie, semantic segmentation), and dense regression (\ie, optical flow estimation).
Source code for this work is available online at \url{http://cs.cmu.edu/~aharley/segaware}. 


\section{Related work}

This work builds on a wide range of research topics. The first is \textit{metric learning.} The goal of metric learning is to produce features from which one can estimate the similarity between pixels or regions in the input \cite{frome2007learning}. Bromley \etal~\cite{siamese} influentially proposed learning these descriptors in a convolutional network, for signature verification. Subsequent related work has yielded compelling results for tasks such as wide-baseline stereo correspondence \cite{han2015matchnet, ZagoruykoCVPR2015, vzbontar2014computing}, and face verification \cite{chopra2005learning}. Recently, the topic of metric learning has been studied extensively in conjunction with image descriptors, such as SIFT and SID \cite{trulls2013dense, simo2015discriminative,pnnet}, improving the applicability of those descriptors to patch-matching problems. Most prior work in metric learning has been concerned with the task of finding one-to-one correspondences between pixels seen from different viewpoints. 
In contrast, the focus of our work is (as in our prior work \cite{harley2016iclr})
 to bring a given point close to all of the other points that lie in the same object. This requires a higher degree of invariance than before -- not only to rotation, scale, and partial occlusion, but also to the interior appearance details of objects. 
Concurrent work has targeted a similar goal, for body joints \cite{newell2016associative} and instance segmentation \cite{fathi2017semantic}. We refer to the features that produce these invariances as \textit{embeddings}, as they embed pixels into a space where the quality of correspondences can be measured as a distance. 

The embeddings in our work are used to generate local \textit{attention masks} to obtain \textit{segmentation-aware} feature maps. The resulting features are meant to capture the appearance of the foreground (relative to a given point), while being invariant to changes in the background or occlusions. To date, related work has focused on developing handcrafted descriptors that have this property. For instance, soft segmentation masks \cite{ott2009implicit,leordeanu2012efficient} and boundary cues \cite{maire2008using,shi2000normalized} have been used to develop segmentation-aware variants of handcrafted features, like SIFT and HOG, effectively suppressing contributions from pixels likely to come from the background \cite{trulls2013dense,trulls2014segmentation}. 
 More in line with the current paper are recent works that incorporate segmentation cues into CNNs, by sharpening or masking intermediate feature maps with the help of superpixels \cite{dai2015convolutional,gadde2016superpixel}. This technique adds spatial structure to multiple stages of the pipeline. 
In all of these works, the affinities are defined in a handcrafted manner, and are typically pre-computed in a separate process. In contrast, we learn the cues directly from image data, and compute the affinities densely and ``on the fly'' within a CNN. Additionally, we combine the masking filters with arbitrary convolutional filters, allowing any layer (or even all layers) to perform segmentation-aware convolution.

Concurrent work in language modelling \cite{dauphin2016language} and image generation \cite{oord2016conditional} has also emphasized the importance of locally masked (or ``gated'') convolutions. Unlike these works, our approach uniquely makes use of embeddings to measure context relevance, which lends interpretability to the masks, and allows for task-agnostic pre-training. Similar attention mechanisms are being used in visual \cite{lu2016hierarchical} and non-visual \cite{sukhbaatar2015end} question answering tasks. These works use a question to construct a single or a limited sequence of globally-supported attention signals. Instead, we use convolutional embeddings, and efficiently construct local attention masks in ``batch mode'' around the region of any given neuron.   

Another relevant thread of works relates to efforts on mitigating the  low-resolution and spatially-imprecise predictions of CNNs. 
Approaches to counter the spatial imprecision weakness can be grouped into preventions (\ie, methods integrated early in the CNN), and cures (\ie, post-processes). A popular preventative method is atrous convolution (also known as ``dilated" convolution) \cite{chen_deeplab, fisher}, which allows neurons to cover a wider field of view with the same number of parameters. Our approach also adjusts neurons' field of view, but focuses it toward the local foreground, rather than widening it in general. 
The ``cures'' aim to restore resolution or sharpness after it has been lost. 
For example, one effective approach is to add trainable upsampling stages to the network, via ``deconvolution'' layers \cite{noh2015learning,long_shelhamer_fcn}. A complementary approach is to stack features from multiple resolutions near the end of the network, so that the final stages have access to both high-resolution (shallow) features and low-resolution (deep) features \cite{hypercols,zoomout,flownet}. Sharpening can be done outside of the CNN, \eg, using edges found in the image \cite{chen2016semantic,bnf}, or using a dense conditional random field (CRF) \cite{koltun2011efficient,chen_deeplab,fisher}. Recently, the CRF approach has been integrated more closely with the CNN, by framing the CRF as a recurrent network, and chaining it to the backpropagation of the underlying CNN \cite{zheng2015conditional}. We make connections and extensions to CRFs in Section 3.3 and provide comparisons in Section 5.1.

\section{Technical approach}

The following subsections describe the main components of our approach. We begin by learning segmentation cues (\refsec{sec:segcues}). We formulate this as a task of finding ``segmentation embeddings'' for the pixels. This step yields features that allow region similarity to be measured as a distance in feature-space. That is, if two pixels have nearby embeddings, then they likely come from the same region.
We next create soft segmentation masks from the embeddings (\refsec{sec:bilateral}).
Our approach generalizes the bilateral filter \cite{lee1983digital,aurich1995non,smith1997susan,tomasi1998bilateral},
which is a technique for creating adaptive smoothing filters that preserve object boundaries.
Noting that CRFs make heavy use of bilateral filters to sharpen posterior estimates, we next describe how to simplify and improve CRFs using our segmentation-aware masks (\refsec{sec:crfs}).
Finally, in \refsec{sec:convolution} we introduce \emph{segmentation-aware convolution}, where we merge segmentation-aware masks with intermediate convolution operations, giving rise to segmentation-aware networks.  


\subsection{Learning segmentation cues}\label{sec:segcues}

The first goal of our work is to obtain segmentation cues.
In particular, we desire features that can be used to infer -- for each pixel -- what other pixels belong to the same object (or scene segment).

Given an RGB image, $\I$, made up of pixels, $\p \in \R^3$ (\ie, 3D vectors encoding color), we learn an embedding function that maps (\ie, embeds) the pixels into a feature space where semantic similarity between pixels can be measured as a distance \cite{chopra2005learning}. Choosing the dimensionality of that feature space to be $D=64$, we can write the embedding function as $f : \R^3 \mapsto \R^D$, or more specifically, $f(\p) = \e$, where $\e$ is the embedding for pixel $\p$.

\begin{figure}[t]
\begin{center}
   \includegraphics[width=0.9\linewidth]{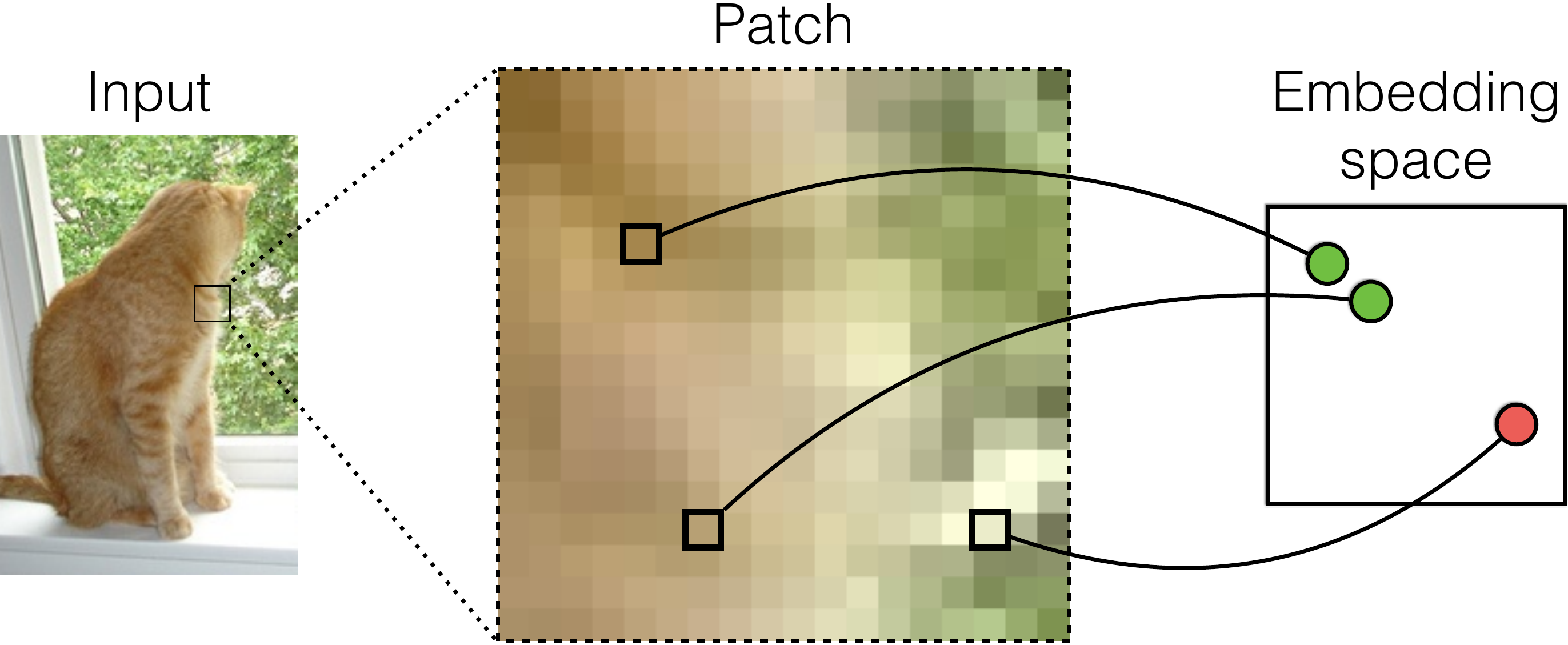}
\end{center}
\vspace{-12pt}
   \caption{Visualization of the goal for pixel embeddings. For any two pixels sampled from the same object, the embeddings should have a small relative distance. For any two pixels sampled from different objects, the embeddings should have a large distance. The embeddings are illustrated in 2D; in principle, they can have any dimensionality.}
\label{fig:embgoal}
\vspace{-6pt}
\end{figure}

Pixel pairs that lie on the same object should produce similar embeddings (\ie, a short distance in feature-space), and pairs from different objects should produce dissimilar embeddings (\ie, a large distance in feature-space). \reffig{fig:embgoal} illustrates this goal with 2D embeddings.
Given semantic category labels for the pixels as training data, we can represent the embedding goal as a loss function over pixel pairs.
For any two pixel indices $i$ and $j$, and corresponding embeddings $\ei, \ej$ and object class labels $\li, \lj$, we can optimize the same-label pairs to have ``near'' embeddings, and the different-label pairs to have ``far'' embeddings. Using $\alpha$ and $\beta$ to denote the ``near'' and ``far'' thresholds, respectively, we can define the pairwise loss as 
\begin{equation} \label{eq:miniloss}
    \lij = 
    \bigg\{
    \begin{array}{lr}
      \text{max} \left( \|\ei - \ej\| - \alpha, 0 \right)  & \text{if } \li = \lj \\
      \text{max} \left( \beta - \|\ei - \ej\|, 0 \right)  & \text{if } \li \neq \lj \\
    \end{array},
\end{equation}
where $\|\boldsymbol{\cdot}\|$ denotes a vector norm. 
We find that embeddings learned from $L^1$ and $L^2$ norms are similar, but $L^1$-based embeddings 
are less vulnerable to exploding gradients. For thresholds, we use $\alpha = 0.5$, and $\beta = 2$. In practice, the specific values of $\alpha$ and $\beta$ are unimportant, so long as $\alpha \leq \beta$ and the remainder of the network can learn to compensate for the scale of the resulting embeddings, \eg, through $\lambda$ in upcoming \refeq{eq:mask}.

To quantify the overall quality of the embedding function, we simply sum the pairwise losses (\refeq{eq:miniloss}) across the image. Although for an image with $N$ pixels there are $N^2$ pairs to evaluate, we find it is effective to simply sample pairs from a neighborhood around each pixel, as in
\begin{equation} \label{eq:loss}
\pazocal{L} = \sum_{i \in N} \sum_{j \in \neigh{i}} \lij,
\end{equation}
where $j \in \neigh{i}$ iterates over the spatial neighbors of index $i$.
In practice, we use three overlapping $3 \times 3$ neighborhoods, with atrous factors \cite{chen_deeplab} of 1, 2, and 5. 
We train a fully-convolutional CNN to minimize this loss through stochastic gradient descent. The network design is detailed in \refsec{sec:arch}.

\begin{figure}[t]
\begin{center}
   \includegraphics[width=1.0\linewidth]{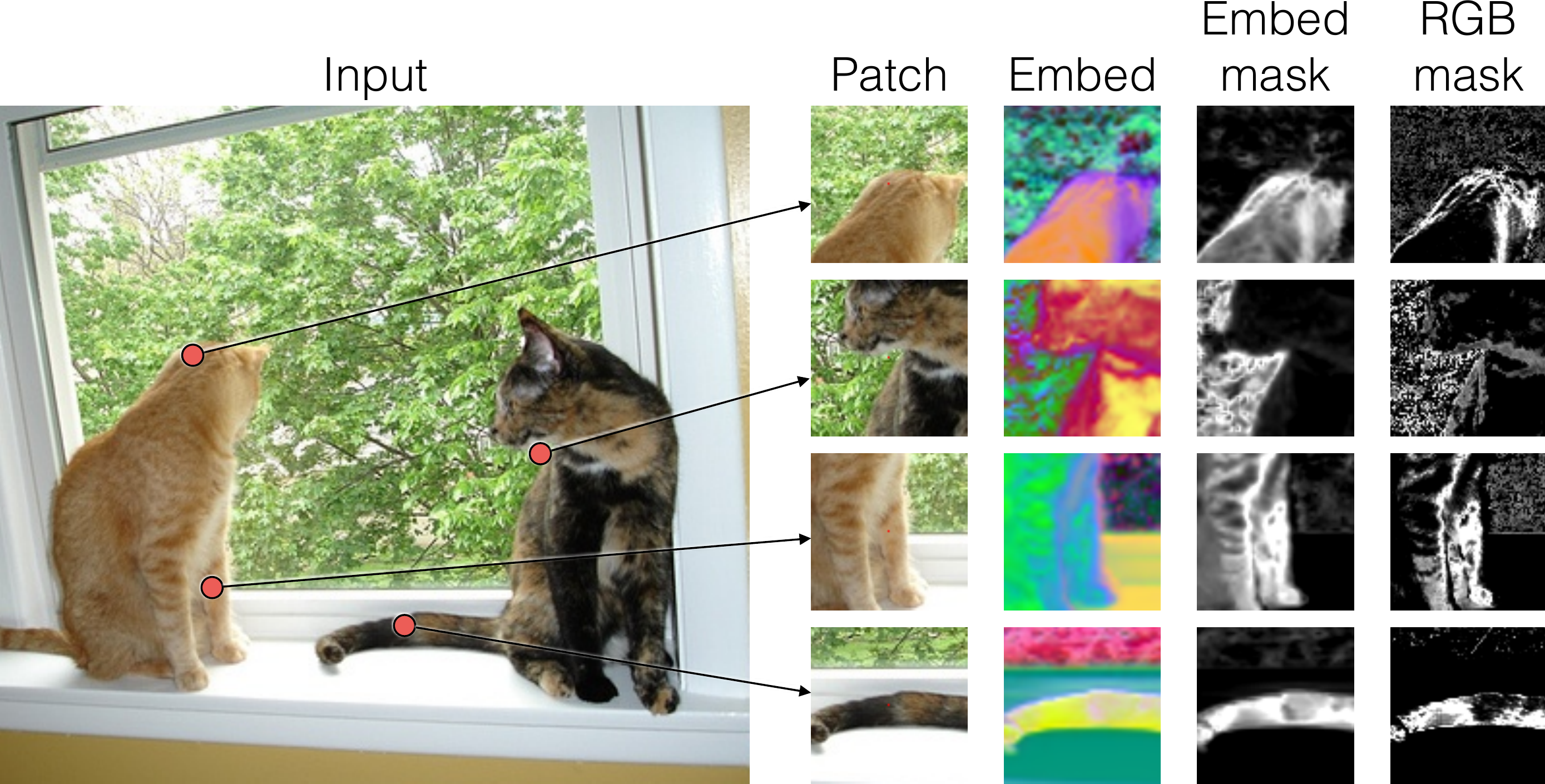}
\end{center}
\vspace{-12pt}
   \caption{Embeddings and local masks are computed densely for input images. For four locations in the image shown on the left, the figure shows (left-to-right) the extracted patch, the embeddings (compressed to three dimensions by PCA for visualization), the embedding-based mask, and the mask generated by color distance.}
\label{fig:rgbvs}
\vspace{-12pt}
\end{figure}

\subsection{Segmentation-aware bilateral filtering}\label{sec:bilateral}
The distance between the embedding at one index, $\ei$, and any other embedding, $\ej$, provides a magnitude indicating whether or not $i$ and $j$ fall on the same object. We can convert these magnitudes into (unnormalized) probabilities, using the exponential distribution:
\begin{equation} \label{eq:mask}
\mij = \exp ( - \lambda \| \ei - \ej \| ),
\end{equation}
where $\lambda$ is a learnable parameter specifying the hardness of this decision, and the notation $\mij$ denotes that $i$ is the reference pixel, and $j$ is the neighbor being considered. In other words, considering all indices $j \in \neigh{i}$, $\mi$ represents a foreground-background segmentation mask, where the central pixel $i$ is defined as the foreground, \ie, $\mii=1$. \reffig{fig:rgbvs} shows examples of the learned segmentation masks (and the intermediate embeddings), and compares them with masks computed from color distances. In general, the learned semantic embeddings successfully generate accurate foreground-background masks, whereas the color-based embeddings are not as reliable.

A first application of these masks is to perform a segmentation-aware smoothing (of pixels, features, or predictions). Given an input feature $\xi$, we can compute a segmentation-aware smoothed result, $\yi$, as follows:
\begin{equation} \label{eq:bilat}
\yi = \frac{\sum_k \x_{i-k} \miik}{\sum_k \miik}, 
\end{equation}
where $k$ is a spatial displacement from index $i$.
Equation~\ref{eq:bilat} has some interesting special cases, which depend on the underlying indexed embeddings $\ej$:
\begin{itemize}
\item if $\ej = 0$, the equation yields the average filter;
\item if $\ej = i$, the equation yields Gaussian smoothing; 
\item if $\ej = (i,\pi)$, where $\pi$ denotes the color vector at $i$, the equation yields bilateral filtering \cite{lee1983digital,aurich1995non,smith1997susan,tomasi1998bilateral}.
\end{itemize}
Since the embeddings are learned in a CNN, \refeq{eq:bilat} represents a generalization of all these cases. 
For comparison, Jampani \etal~\cite{jampani15learning} propose to learn the kernel used in the bilateral filter, but keep the arguments to the similarity measure (\ie, $\ei$) fixed. In our work, by training the network to provide convolutional embeddings, we additionally learn the arguments of the bilateral distance function. 

\begin{figure}[t]
\begin{center}
   \includegraphics[width=1.0\linewidth]{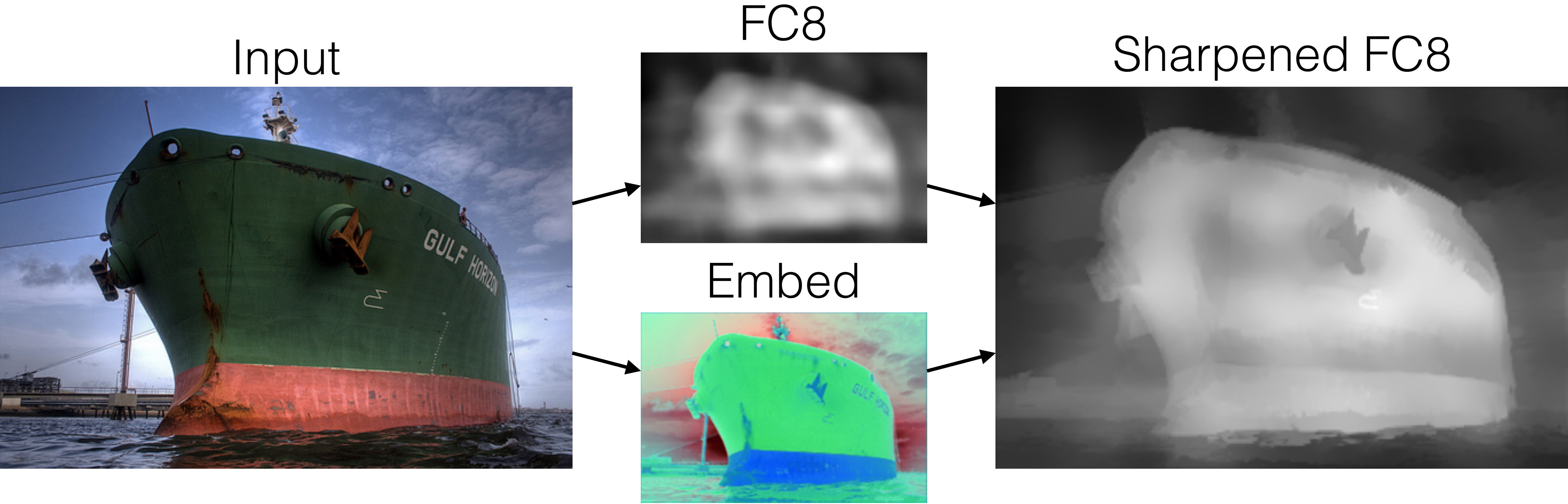}
\end{center}
\vspace{-12pt}
\caption{Segmentation-aware bilateral filtering. Given an input image (left), a CNN typically produces a smooth prediction map (middle top). Using learned per-pixel embeddings (middle bottom), we adaptively smooth the FC8 feature map with our segmentation-aware bilateral filter (right).}
\label{fig:bilat}
\vspace{-12pt}
\end{figure}

When the embeddings are integrated into a larger network that uses them for filtering, the embedding loss function (\refeq{eq:loss}) is no longer necessary. Since all of the terms in the filter function (\refeq{eq:bilat}) are differentiable, the global objective (\eg, classification accuracy) can be used to tune not only the input terms, $\xi$, but also the mask terms, $\mij$, and their arguments, $\ej$. Therefore, the embeddings can be learned end-to-end in the network when used to create masks. In our work, we first train the embeddings with a dedicated loss, then fine-tune them in the larger pipeline in which they are used for masks.

\reffig{fig:bilat} shows an example of how segmentation-aware bilateral filtering sharpens FC8 predictions in practice.

\subsection{Segmentation-aware CRFs}\label{sec:crfs}

Segmentation-aware bilateral filtering can be used to improve CRFs. As discussed earlier, dense CRFs \cite{koltun2011efficient} are effective at sharpening the prediction maps produced by CNNs \cite{chen_deeplab,zheng2015conditional}.

These models optimize a Gibbs energy given by \begin{equation}
E(\textbf{x}) = \sum_i \psi_u (\xi) + \sum_i \sum_{j \leq i} \psi_p (\xi, \xj), \end{equation}where $i$ ranges over all pixel indices in the image. In semantic segmentation, the unary term $\psi_u$ is typically chosen to be the negative log probability provided by a CNN trained for per-pixel classification. The pairwise potentials take the form $\psi_p (\xi, \xj) = \mu(\xi,\xj) k(\textbf{f}_i, \textbf{f}_j)$, where $\mu$ is a label compatibility function (\eg, the Potts model), and $k(\textbf{f}_i, \textbf{f}_j)$ is a feature compatibility function. The feature compatibility is composed of an appearance term (a bilateral filter), and a smoothness term (an averaging filter), in the form
\begin{equation}\label{eq:pairwise}
\begin{split}
k(\textbf{f}_i, \textbf{f}_j) &= w^1 \exp \left( - \frac{\|i - j\|^2}{2 \theta^2_\alpha}  - \frac{\| \pi - \pj \|^2}{2 \theta^2_\beta} \right) \\
& + w^{2} \exp \left( - \frac{\|i - j\|^2}{2 \theta^2_\gamma} \right),
\end{split}
\end{equation}where the $w^{k}$ are weights on the two terms. Combined with the label compatibility function, the appearance term adds a penalty if a pair of pixels are assigned the same label but have dissimilar colors. 
To be effective, these filtering operations are carried out with extremely wide filters (\eg, the size of the image), which necessitates using a data structure called a permutohedral lattice \cite{adams2010fast}.

Motivated by our earlier observation that learned embeddings are a stronger semantic similarity signal than color (see Fig.~\ref{fig:rgbvs}), we replace the color vector $\pi$ in \refeq{eq:pairwise} with the learned embedding vector $\ei$. The permutohedral lattice would be inefficient for such a high-dimensional filter, but we find that the signal provided by the embeddings is rich enough that we can use small filters (\eg, $13 \times 13$), and achieve the same (or better) performance. This allows us to implement the entire CRF with standard convolution operators, reduce computation time by half, and backpropagate through the CRF into the embeddings.

\subsection{Segmentation-aware convolution}\label{sec:convolution}

The bilateral filter in \refeq{eq:bilat} is similar in form to convolution, but with a non-linear sharpening mask instead of a learned task-specific filter. In this case, 
we can have the benefits of both, by inserting the learned convolution filter, $\t$, into the equation:
\begin{equation} \label{eq:segaware}
\yi = \frac{\sum_k \x_{i-k} \miik \t_k}{\sum_k \miik}.
\end{equation}
This is a non-linear convolution: the input signal is multiplied pointwise by the normalized local mask before forming the inner product with the learned filter. If the learned filter $\t_i$ is all ones, we have the same bilateral filter as in \refeq{eq:bilat}; if the embedding-based segmentation mask $\m_i$ is all ones, we have standard convolution. Since the masks in this context encode segmentation cues, we refer to \refeq{eq:segaware} as segmentation-aware convolution.


The mask acts as an applicability function for the filter, which makes segmentation-aware convolution a special case of normalized convolution \cite{knutsson1993normalized}. The idea of normalized convolution is to ``focus'' the convolution operator on the part of the input that truly describes the input signal, avoiding the interpolation of noise or missing information. In this case, ``noise'' corresponds to information coming from regions other than the one to which index $i$ belongs. 




Any convolution filter can be made segmentation-aware. The advantage of segmentation awareness depends on the filter. For instance, a center-surround filter might be rendered useless by the effect of the mask (since it would block the input from the ``surround''), whereas a filter selective to a particular shape might benefit from invariance to context. The basic intuition is that the information masked out needs to be distracting rather than helping; realizing this in practice requires learning the masking functions. In our work, we use backpropagation to learn both the arguments and the softness of each layer's masking operation, \ie, both $\ei$ and $\lambda$ in \refeq{eq:mask}. Note that the network can always fall back to a standard CNN by simply learning a setting of $\lambda=0$.

\section{Implementation details}\label{sec:arch}

This section first describes how the basic ideas of the technical approach are integrated in a CNN architecture, and then provides details on how the individual components are implemented efficiently as convolution-like layers.

\begin{figure*}[t]
\begin{center}
   \includegraphics[width=1.0\linewidth]{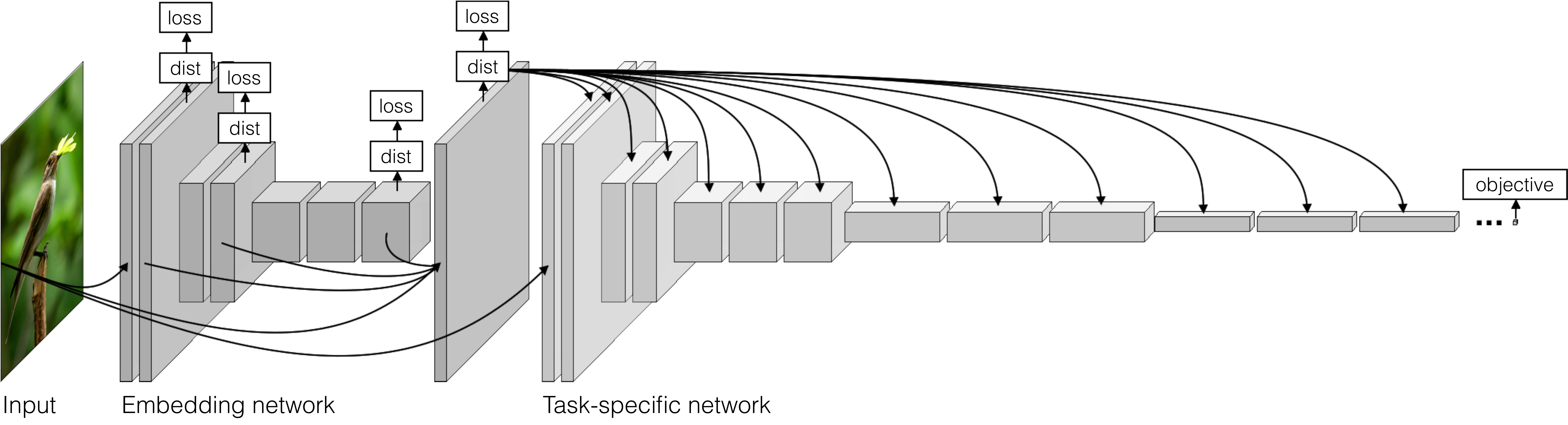}
\end{center}
\vspace{-12pt}
   \caption{General schematic for our segmentation-aware CNN. The first part is an embedding network, which is guided to compute embedding-like representations at multiple scales, and constructs a final embedding as a weighted sum of the intermediate embeddings. The loss on these layers operates on pairwise distances computed from the embeddings. These same distances are then used to construct local attention masks, that intercept the convolutions in a task-specific network. The final objective  backpropagates through both networks, fine-tuning the embeddings for the task.}
\label{fig:schematic}
\vspace{-12pt}
\end{figure*}

\subsection{Network architecture}

Any convolutional network can be made segmentation-aware. In our work, the technique for achieving this modification involves generating embeddings with a dedicated ``embedding network'', then using masks computed from those embeddings to modify the convolutions of a given task-specific network. This implementation strategy is illustrated in \reffig{fig:schematic}.


The embedding network has the following architecture. The first seven layers share the design of the earliest convolution layers in VGG-16 \cite{Chatfield14}, and are initialized with that network's (object recognition-trained) weights. There is a subsampling layer after the second convolution layer and also after the fourth convolution layer, so the network captures information at three different scales. The final output from each scale is sent to a pairwise distance computation (detailed in \refsec{sec:eff}) followed by a loss (as in \refeq{eq:miniloss}), so that each scale develops embedding-like representations. The outputs from the intermediate embedding layers are then upsampled to a common resolution, concatenated, and sent to a convolution layer with $1 \times 1$ filters. This layer learns a weighted average of the intermediate embeddings, and creates the final embedding for each pixel. 

The idea of using a loss at intermediate layers is inspired by Xie and Tu~\cite{xie15hed}, who used this strategy to learn boundary cues in a CNN. The motivation behind this strategy is to provide early layers a stronger signal of the network's end goal, reducing the burden on backpropagation to carry the signal through multiple layers \cite{lee2015deeply}. 


The final embeddings are used to create masks in the task-specific network. The lightest usage of these masks involves performing segmentation-aware bilateral filtering on the network's final layer outputs; this achieves the sharpening effect illustrated in \reffig{fig:bilat}. The most intrusive usage of the masks involves converting all convolutions into segmentation-aware convolutions. Even in this case, however, the masks can be inserted with no detrimental effect (\ie, by initializing with $\lambda=0$ in \refeq{eq:mask}), allowing the network to learn whether or not (and at what layer) to activate the masks. Additionally, if the target task has discrete output labels, as in the case of semantic segmentation, a segmentation-aware CRF can be attached to the end of the network to sharpen the final output predictions. 

\subsection{Efficient convolutional implementation details}\label{sec:eff}

We reduce all steps of the pipeline to matrix multiplications, making the approach very efficient on GPUs. We achieve this by casting the mask creation (\ie, pairwise embedding distance computation) as a convolution-like operation, and implementing it in exactly the way Caffe \cite{jia2014caffe} realizes convolution: via an image-to-column transformation, followed by matrix multiplication.

More precisely, the distance computation works as follows. For every position $i$ in the feature-map provided by the layer below, a patch of features is extracted from the neighborhood $j \in \neigh{i}$, and distances are computed between the central feature and its neighbors. These distances are arranged into a row vector of length $K$, where $K$ is the spatial dimensionality of the patch. This process turns an $H \times W$ feature-map into an $H \cdot W \times K$ matrix, where each element in the $K$ dimension holds a distance relating that pixel to the central pixel at that spatial index. 

To convert the distances into masks, the $H \cdot W \times K$ matrix is passed through an exponential function with a specified hardness, $\lambda$. This operation realizes the mask term (\refeq{eq:mask}). In our work, the hardness of the exponential is learned as a parameter of the CNN.

To perform the actual masking, the input to be masked is simply processed by an image-to-column transformation (producing another $H \cdot W \times K$ matrix), then multiplied pointwise with the normalized mask matrix. 
From that product, segmentation-aware bilateral filtering is merely a matter of summing across the $K$ dimension, producing an $H \cdot W \times 1$ matrix that can be reshaped into dimensions $H \times W$. 
Segmentation-aware convolution (\refeq{eq:segaware}) simply requires multiplying the $H \cdot W \times K$ masked values with a $K \times F$ matrix of weights, where $F$ is the number of convolution filters. The result of this multiplication can be reshaped into $F$ different $H \times W$ feature maps.



\section{Evaluation}
We evaluate on two different dense prediction tasks: semantic segmentation, and optical flow estimation. The goal of the experiments is to minimally modify strong baseline networks, and examine the effects of instilling various levels of ``segmentation awareness''. 

\subsection{Semantic segmentation}
Semantic segmentation is evaluated on the PASCAL VOC 2012 challenge \cite{pascal-voc-2012}, augmented with additional images from Hariharan \etal~\cite{hariharan2011semantic}. 
Experiments are carried out with two different baseline networks, ``DeepLab'' \cite{chen_deeplab} and ``DeepLabV2''  \cite{deeplabv2}. DeepLab is a fully-convolutional version of VGG-16 \cite{Chatfield14}, using atrous convolution in some layers to reduce downsampling. DeepLabV2 is a fully-convolutional version of a 101-layer residual network (ResNet) \cite{he2016deep}, modified with atrous spatial pyramid pooling and multi-scale input processing. Both networks are initialized with weights learned on ImageNet \cite{ILSVRC15}, then trained on the Microsoft COCO training and validation sets 
\cite{coco}, and finally fine-tuned on the PASCAL images \cite{pascal-voc-2012,hariharan2011semantic}. 

To replace the densely connected CRF used in the original works \cite{chen_deeplab, deeplabv2}, we attach a very sparse segmentation-aware CRF. We select the hyperparameters of the segmentation-aware CRF via cross validation on a small subset of the validation set, arriving at a $ 13 \times 13$ bilateral filter with an atrous factor of 9, a $5 \times 5$ spatial filter, and 2 meanfield iterations for both training and testing.

We carry out the main set of experiments with DeepLab on the VOC validation set, investigating the piecewise addition of various segmentation-aware components.  A summary of the results is presented in Table~\ref{table:vocval}. The first result is that using learned embeddings to mask the output of DeepLab approximately provides a 0.6\% improvement in mean intersection-over-union (IOU) accuracy. This is achieved with a single application of a $9 \times 9$ bilateral-like filter on the FC8 outputs produced by DeepLab. 

\begin{table}
\centering
  \caption{PASCAL VOC 2012 validation results for the various considered approaches, compared against the baseline. All methods use DeepLab as the base network; ``BF'' means bilateral filter; ``SegAware'' means segmentation-aware.}
    \vspace{-12pt}
  \label{table:vocval}
  \begin{center}
  \begin{tabular}{ll}
    \toprule
    Method     						& IOU (\%)\\
    \midrule
    DeepLab						& 66.33\\
    \ldots + CRF						& 67.60\\
\hline
    \ldots + $9 \times 9$ SegAware BF	& 66.98 \\
    \ldots + $9 \times 9$ SegAware BF $\times 2$ 	& 67.36 \\
    \ldots + $9 \times 9$ SegAware BF $\times 4$ 	& 67.68 \\
\hline
    \ldots with FC6 SegAware								& 67.40	\\
    \ldots with all layers SegAware						& 67.94	\\
\hline
    \ldots with all layers SegAware + $9 \times 9$ BF 			& 68.00 \\
    \ldots with all layers SegAware + $7 \times 7$ BF $ \times 2$ & 68.57 \\
	\ldots with all layers SegAware + $5 \times 5$ BF $ \times 4$ 		& 68.52 \\
\hline
	\ldots with all layers and CRF SegAware		& \textbf{69.01} 		\\
	\bottomrule
	  \end{tabular}
  \end{center}
  \vspace{-6pt}
\end{table}


Once the embeddings and masks are computed, it is straightforward to run the masking process repeatedly. Applying the process multiple times improves performance by strengthening the contribution from similar neighbors in the radius, and also by allowing information from a wider radius to contribute to each prediction. Applying the bilateral filter four times increases the gain in IOU accuracy to 1.3\%. This is at the cost of approximately 500 ms of additional computation time. A dense CRF yields slightly worse performance, at approximately half the speed (1 second).

\begin{figure}[t]
\begin{center}
   \includegraphics[width=1.0\linewidth]{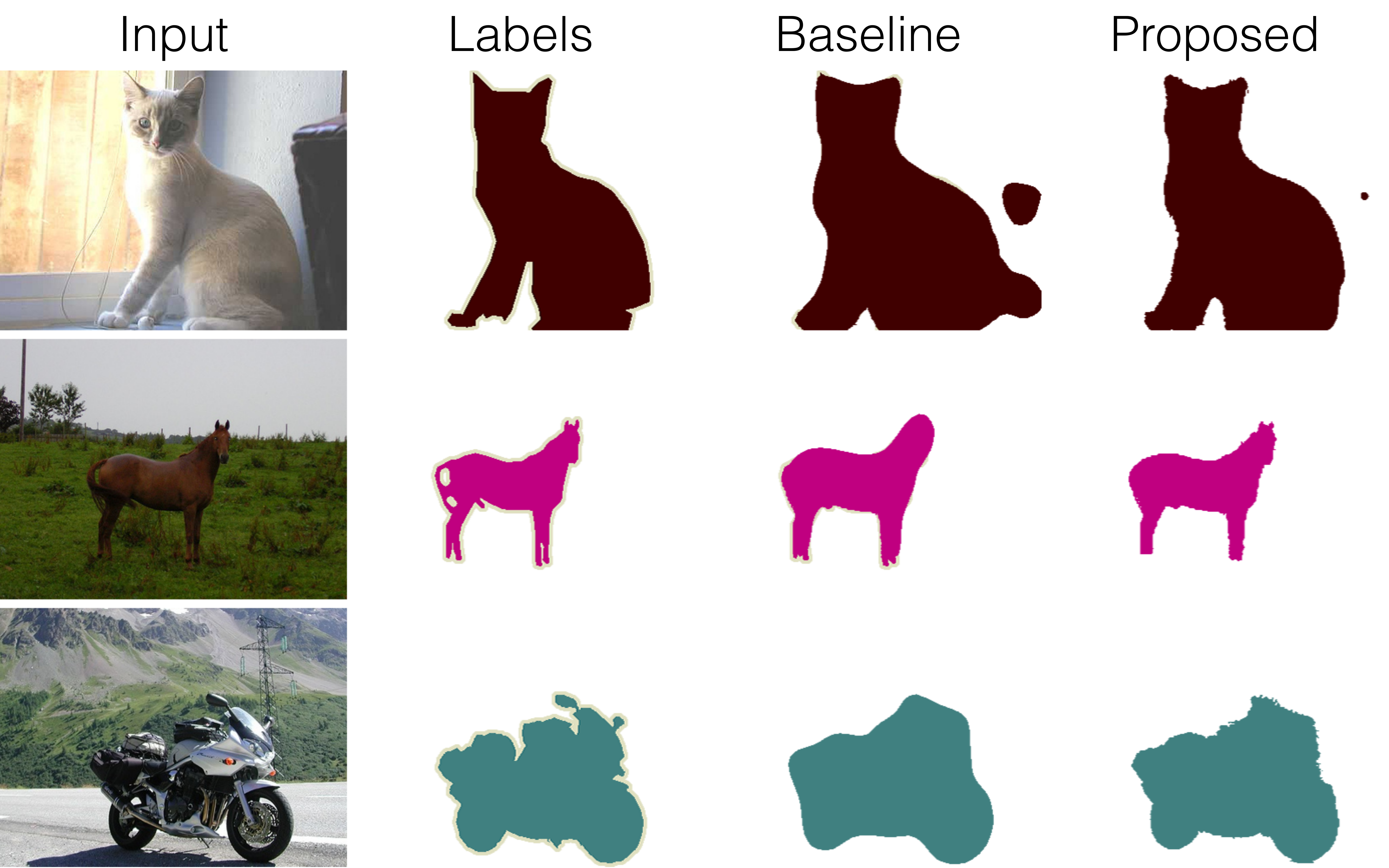}
\end{center}
\vspace{-6pt}
   \caption{Visualizations of semantic segmentations produced by DeepLab and its segmentation-aware variant on the PASCAL VOC 2012 validation set.}
\label{fig:niceoutputs}
\vspace{-6pt}
\end{figure}

Segmentation-aware convolution provides similar improvements, at less computational cost. Simply making the FC6 layer segmentation-aware produces an improvement of approximately 1\% to IOU accuracy, at a cost of +100 ms, while making all layers segmentation-aware improves accuracy by 1.6\%, at a cost of just +200 ms. 

\begin{figure}[t]
  \centering
  \subfloat
  {\includegraphics[width=0.36\linewidth]{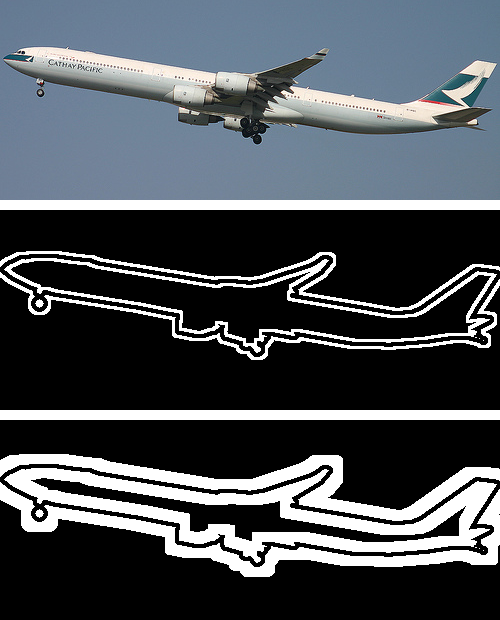}\label{fig:f1}}
  \hfill
  \subfloat
  {\includegraphics[width=0.58\linewidth]{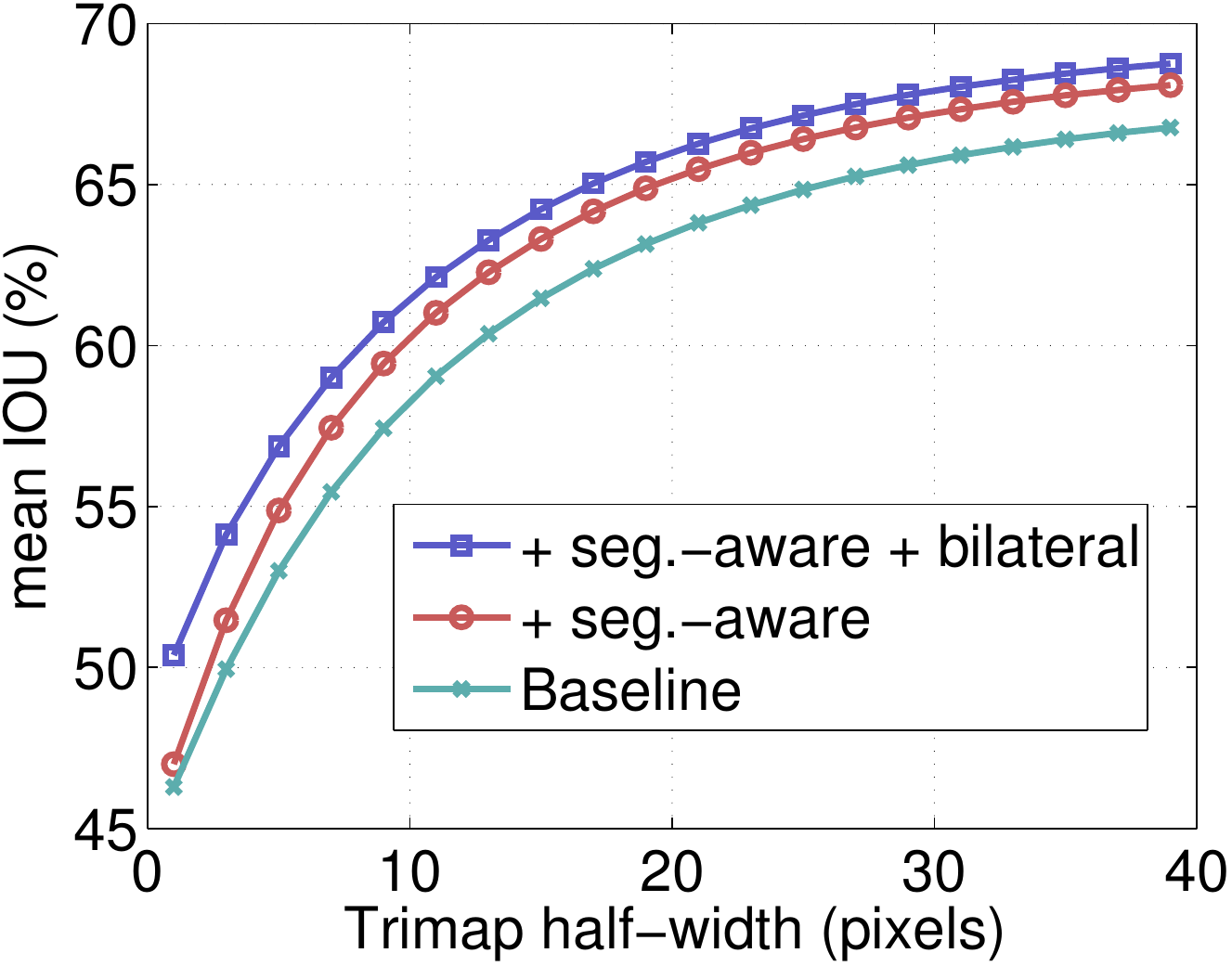}\label{fig:f2}}
  \caption{Performance near object boundaries (``trimaps''). Example trimaps are visualized (in white) for the image in the top left; the trimap of half-width three is shown in the middle left, and the trimap of half-width ten is shown on the bottom left. Mean IOU performance of the baseline and two segmentation-aware variants are plotted (right) for trimap half-widths 1 to 40.}
\label{fig:trimap}
\vspace{-6pt}
\end{figure}

To examine where the gains are taking place, we compute each method's accuracy within ``trimaps'' that extend from the objects' boundaries. A trimap is a narrow band (of a specified half-width) that surrounds a boundary on either side; measuring accuracy exclusively within this band can help separate within-object accuracy from on-boundary accuracy \cite{chen_deeplab}. \reffig{fig:trimap} (left) shows examples of trimaps, and (right) plots accuracies as a function of trimap width. The results show that segmentation-aware convolution offers its main improvement slightly away from the boundaries (\ie, beyond 10 pixels), while bilateral filtering offers its largest improvement very near the boundary (\ie, within 5 pixels).

Combining segmentation-aware convolution with bilateral filtering pushes the gains to 2.2\%. 
Finally, adding a segmentation-aware CRF to the pipeline increases IOU accuracy by an additional 0.5\%, bringing the overall gain to approximately 2.7\% over the DeepLab baseline. 

We evaluate the ``all components'' approach on the VOC test server, with both DeepLab and DeepLabV2. Results are summarized in Table~\ref{table:voctest}. The improvement over DeepLab is 2\%, which is noticeable in visualizations of the results, as shown in \reffig{fig:niceoutputs}. DeepLabV2 performs approximately 10 points higher than DeepLab; we exceed this improvement by approximately 0.8\%. The segmentation-aware modifications perform equally well (0.1\% superior) to dense CRF post-processing, despite being simpler (using only a sparse CRF, and replacing the permutohedral lattice with basic convolution), and twice as fast (0.5s rather than 1s). 

\begin{table}
\centering
  \caption{PASCAL VOC 2012 test results.}
    \vspace{-6pt}
  \label{table:voctest}
  \begin{center}
  \begin{tabular}{ll}
    \toprule
    Method     						& IOU (\%) 		\\
    \midrule
    DeepLab 						& 67.0 	\\
    DeepLab+CRF						& 68.2 	\\
    SegAware DeepLab                & 69.0\\
    DeepLabV2 						& 79.0	\\
    DeepLabV2+CRF 					& 79.7	\\
    SegAware DeepLabV2              & \textbf{79.8} \\
    \bottomrule
  \end{tabular}
  \end{center}
  \vspace{-6pt}
\end{table}

\subsection{Optical flow}
We evaluate optical flow on the recently introduced FlyingChairs \cite{flownet} dataset. The baseline network for this experiment is the ``FlowNetSimple'' model from Dosovitskiy \etal~\cite{flownet}. This is a fully-convolutional network, with a contractive part that reduces the resolution of the input by a factor of 64, and an expansionary part (with skip connections) that restores the resolution to quarter-size. 

\begin{figure}[t]
\begin{center}
   \includegraphics[width=1.0\linewidth]{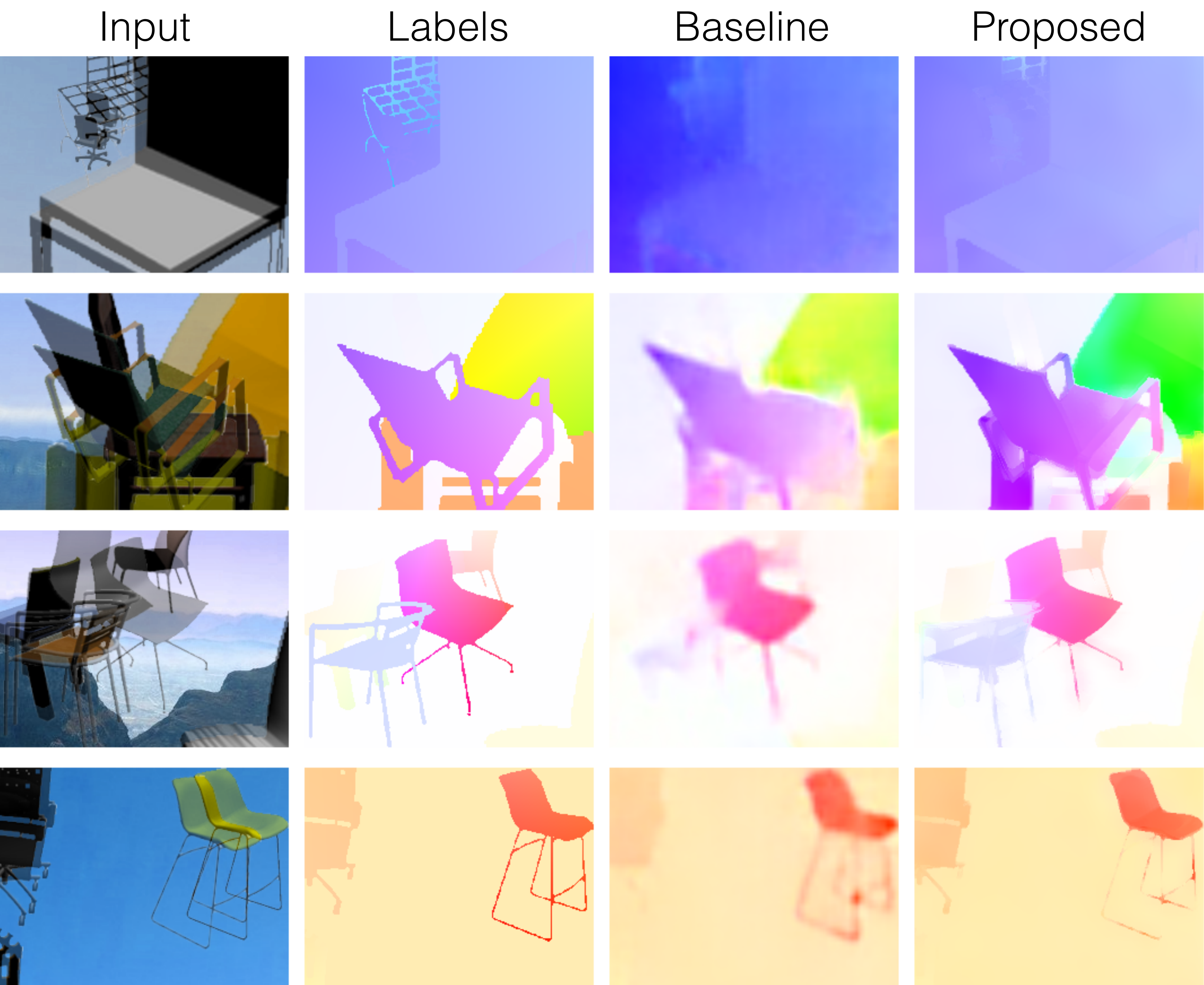}
\end{center}
\vspace{-6pt}
   \caption{Visualizations of optical flow produced by FlowNet and its segmentation-aware variant on the FlyingChairs test set: segmentation-awareness yields much sharper results than the baseline.}
\label{fig:flows}
  \vspace{-6pt}
\end{figure}

In this context, we find that relatively minor segmentation-aware modifications yield substantial gains in accuracy. Using embeddings pre-trained on PASCAL VOC, we make the final prediction layer segmentation-aware, and add $9 \times 9$ bilateral filtering to the end of the network. This reduces the average end-point error (aEPE) from 2.78 to 2.26 (an 18\% reduction in error), and reduces average angular error by approximately 6 degrees, from 15.58 to 9.54. We achieve these gains without the aggressive data augmentation techniques pursued by Dosovitskiy \etal~\cite{flownet}. Table~\ref{table:flowval} lists these results in the context of some related work in this domain, demonstrating that the gain is fairly substantial. FlowNetCorr \cite{flownet} achieves a better error, but it effectively doubles the network size and runtime, whereas our method only adds a shallow set of embedding layers. As shown in \reffig{fig:flows}, a qualitative improvement to the flow fields is easily discernable, especially near object boundaries. Note that the performance of prior FlowNet architectures diminishes with the application of variational refinement \cite{flownet}, likely because this step was not integrated in the training process. The filtering methods of this work, however, are easily integrated into backpropagation.


\begin{table}
\centering
  \caption{FlyingChairs test results.}
      \vspace{-12pt}
  \label{table:flowval}
  \begin{center}
  \begin{tabular}{lll}
    \toprule
    Method     						& aEPE & aAE 	\\
    \midrule
    SPyNet \cite{spynet} & 2.63 & - \\
    EpicFlow \cite{revaud2015epicflow} 					& 2.94	 & -\\
    DeepFlow \cite{weinzaepfel2013deepflow}						& 3.53 	 & -\\
    LDOF \cite{brox2011large}						& 3.47 	 & -\\
    FlowNetSimple \cite{flownet}						& 2.78	 & 15.58\\
    FlowNetSimple + variational \cite{flownet}						& 2.86	 & - \\
    FlowNetCorr	 \cite{flownet}					& \textbf{2.19}	 & - \\
    FlowNetCorr + variational \cite{flownet}					& 2.61	 & - \\
    SegAware FlowNetSimple 	& 2.36 & \textbf{9.54} \\
	\bottomrule
  \end{tabular}
  \end{center}
  \vspace{-12pt}
\end{table}

\section{Conclusion}
This work introduces Segmentation-Aware Convolutional Networks, a direct generalization of standard CNNs that allows us to seamlessly accommodate segmentation information throughout a deep architecture. Our approach avoids feature blurring before it happens, rather than fixing it post-hoc. The full architecture can be trained end-to-end. We have shown that this allows us to directly compete with segmentation-specific structured prediction algorithms, while easily extending to continuous prediction tasks, such as optical flow estimation, that currently have no remedy for blurred responses.


{\small
\bibliographystyle{ieee}
\bibliography{bibref_definitions_short,refs}

\begin{thebibliography}{10}\itemsep=-1pt

\bibitem{adams2010fast}
A.~Adams, J.~Baek, and M.~A. Davis.
\newblock Fast high-dimensional filtering using the permutohedral lattice.
\newblock {\em Computer Graphics Forum}, 29(2):753--762, 2010.

\bibitem{aurich1995non}
V.~Aurich and J.~Weule.
\newblock Non-linear gaussian filters performing edge preserving diffusion.
\newblock In {\em {Proceedings of the DAGM Symposium}}, pages 538--545, 1995.

\bibitem{pnnet}
V.~Balntas, E.~Johns, L.~Tang, and K.~Mikolajczyk.
\newblock {PN}-{N}et: {C}onjoined triple deep network for learning local image
  descriptors.
\newblock {\em arXiv:1601.05030}, 2016.

\bibitem{bnf}
G.~Bertasius, J.~Shi, and L.~Torresani.
\newblock Semantic segmentation with boundary neural fields.
\newblock In {\em CVPR}, 2016.

\bibitem{siamese}
J.~Bromley, I.~Guyon, Y.~Lecun, E.~Sackinger, and R.~Shah.
\newblock Signature verification using a ``siamese'' time delay neural network.
\newblock In {\em NIPS}, 1994.

\bibitem{brox2011large}
T.~Brox and J.~Malik.
\newblock Large displacement optical flow: {D}escriptor matching in variational
  motion estimation.
\newblock {\em PAMI}, 33(3):500--513, 2011.

\bibitem{Chatfield14}
K.~Chatfield, K.~Simonyan, A.~Vedaldi, and A.~Zisserman.
\newblock Return of the devil in the details: {D}elving deep into convolutional
  nets.
\newblock In {\em BMVC}, 2014.

\bibitem{chen2016semantic}
L.-C. Chen, J.~T. Barron, G.~Papandreou, K.~Murphy, and A.~L. Yuille.
\newblock Semantic image segmentation with task-specific edge detection using
  {CNN}s and a discriminatively trained domain transform.
\newblock In {\em CVPR}, 2016.

\bibitem{chen_deeplab}
L.-C. Chen, G.~Papandreou, I.~Kokkinos, K.~Murphy, and A.~L. Yuille.
\newblock Semantic image segmentation with deep convolutional nets and fully
  connected {CRF}s.
\newblock In {\em ICLR}, 2015.

\bibitem{deeplabv2}
L.-C. Chen, G.~Papandreou, I.~Kokkinos, K.~Murphy, and A.~L. Yuille.
\newblock Deep{L}ab: {S}emantic image segmentation with deep convolutional
  nets, atrous convolution, and fully connected {CRF}s.
\newblock {\em PAMI}, 2016.

\bibitem{chopra2005learning}
S.~Chopra, R.~Hadsell, and Y.~LeCun.
\newblock Learning a similarity metric discriminatively, with application to
  face verification.
\newblock In {\em CVPR}, 2005.

\bibitem{dai2015convolutional}
J.~Dai, K.~He, and J.~Sun.
\newblock Convolutional feature masking for joint object and stuff
  segmentation.
\newblock In {\em CVPR}, 2015.

\bibitem{dauphin2016language}
Y.~N. Dauphin, A.~Fan, M.~Auli, and D.~Grangier.
\newblock Language modeling with gated convolutional networks.
\newblock {\em arXiv:1612.08083}, 2016.

\bibitem{flownet}
A.~Dosovitskiy, P.~Fischer, E.~Ilg, P.~H{\"a}usser, C.~Haz{\i}rba{\c{s}},
  V.~Golkov, P.~van~der Smagt, D.~Cremers, and T.~Brox.
\newblock Flow{N}et: {L}earning optical flow with convolutional networks.
\newblock In {\em ICCV}, 2015.

\bibitem{eigen2014depth}
D.~Eigen, C.~Puhrsch, and R.~Fergus.
\newblock Depth map prediction from a single image using a multi-scale deep
  network.
\newblock In {\em NIPS}, 2014.

\bibitem{pascal-voc-2012}
M.~Everingham, L.~Van-Gool, C.~K.~I. Williams, J.~Winn, and A.~Zisserman.
\newblock The {PASCAL} {V}isual {O}bject {C}lasses {C}hallenge 2012 {(VOC2012)}
  {R}esults.
\newblock
  http://www.pascal-network.org/challenges/VOC/voc2012/workshop/index.html,
  2012.

\bibitem{fathi2017semantic}
A.~Fathi, Z.~Wojna, V.~Rathod, P.~Wang, H.~O. Song, S.~Guadarrama, and K.~P.
  Murphy.
\newblock Semantic instance segmentation via deep metric learning.
\newblock {\em arXiv:1703.10277}, 2017.

\bibitem{frome2007learning}
A.~Frome, Y.~Singer, F.~Sha, and J.~Malik.
\newblock Learning globally-consistent local distance functions for shape-based
  image retrieval and classification.
\newblock In {\em ICCV}, 2007.

\bibitem{gadde2016superpixel}
R.~Gadde, V.~Jampani, M.~Kiefel, and P.~V. Gehler.
\newblock Superpixel convolutional networks using bilateral inceptions.
\newblock In {\em ECCV}, 2016.

\bibitem{han2015matchnet}
X.~Han, T.~Leung, Y.~Jia, R.~Sukthankar, and A.~Berg.
\newblock Match{N}et: {U}nifying feature and metric learning for patch-based
  matching.
\newblock In {\em CVPR}, 2015.

\bibitem{hariharan2011semantic}
B.~Hariharan, P.~Arbel{\'a}ez, L.~Bourdev, S.~Maji, and J.~Malik.
\newblock Semantic contours from inverse detectors.
\newblock In {\em ICCV}, 2011.

\bibitem{hypercols}
B.~Hariharan, P.~Arbel{\'a}ez, R.~Girshick, and J.~Malik.
\newblock Hypercolumns for object segmentation and fine-grained localization.
\newblock In {\em CVPR}, 2015.

\bibitem{harley2016iclr}
A.~W. Harley, K.~G. Derpanis, and I.~Kokkinos.
\newblock Learning dense convolutional embeddings for semantic segmentation.
\newblock In {\em ICLR}, 2016.

\bibitem{he2016deep}
K.~He, X.~Zhang, S.~Ren, and J.~Sun.
\newblock Deep residual learning for image recognition.
\newblock In {\em CVPR}, 2016.

\bibitem{jampani15learning}
V.~Jampani, M.~Kiefel, and P.~V. Gehler.
\newblock Learning sparse high dimensional filters: {I}mage filtering, dense
  {CRF}s and bilateral neural networks.
\newblock In {\em CVPR}, 2016.

\bibitem{jia2014caffe}
Y.~Jia, E.~Shelhamer, J.~Donahue, S.~Karayev, J.~Long, R.~Girshick,
  S.~Guadarrama, and T.~Darrell.
\newblock Caffe: {C}onvolutional architecture for fast feature embedding.
\newblock In {\em ACM-MM}, pages 675--678, 2014.

\bibitem{knutsson1993normalized}
H.~Knutsson and C.-F. Westin.
\newblock Normalized and differential convolution.
\newblock In {\em CVPR}, 1993.

\bibitem{koltun2011efficient}
P.~Kr\"{a}henb\"{u}hl and V.~Koltun.
\newblock Efficient inference in fully connected {CRF}s with {G}aussian edge
  potentials.
\newblock In {\em NIPS}, 2011.

\bibitem{kriz}
A.~Krizhevsky, I.~Sutskever, and G.~E. Hinton.
\newblock Image{N}et classification with deep convolutional neural networks.
\newblock In {\em NIPS}, 2012.

\bibitem{lee2015deeply}
C.-Y. Lee, S.~Xie, P.~Gallagher, Z.~Zhang, and Z.~Tu.
\newblock Deeply-supervised nets.
\newblock {\em AISTATS}, 2(3):6, 2015.

\bibitem{lee1983digital}
J.-S. Lee.
\newblock Digital image smoothing and the sigma filter.
\newblock {\em CVGIP}, 24(2):255--269, 1983.

\bibitem{leordeanu2012efficient}
M.~Leordeanu, R.~Sukthankar, and C.~Sminchisescu.
\newblock Efficient closed-form solution to generalized boundary detection.
\newblock In {\em ECCV}, pages 516--529, 2012.

\bibitem{coco}
T.-Y. Lin, M.~Maire, S.~Belongie, J.~Hays, P.~Perona, D.~Ramanan,
  P.~Doll{\'a}r, and C.~L. Zitnick.
\newblock Microsoft {COCO}: {C}ommon objects in context.
\newblock In {\em ECCV}, pages 740--755, 2014.

\bibitem{long_shelhamer_fcn}
J.~Long, E.~Shelhamer, and T.~Darrell.
\newblock Fully convolutional networks for semantic segmentation.
\newblock In {\em CVPR}, 2014.

\bibitem{lu2016hierarchical}
J.~Lu, J.~Yang, D.~Batra, and D.~Parikh.
\newblock Hierarchical question-image co-attention for visual question
  answering.
\newblock In {\em NIPS}, pages 289--297, 2016.

\bibitem{maire2008using}
M.~Maire, P.~Arbel{\'a}ez, C.~Fowlkes, and J.~Malik.
\newblock Using contours to detect and localize junctions in natural images.
\newblock In {\em CVPR}, 2008.

\bibitem{zoomout}
M.~Mostajabi, P.~Yadollahpour, and G.~Shakhnarovich.
\newblock Feedforward semantic segmentation with zoom-out features.
\newblock In {\em CVPR}, 2015.

\bibitem{newell2016associative}
A.~Newell and J.~Deng.
\newblock Associative embedding: {E}nd-to-end learning for joint detection and
  grouping.
\newblock {\em arXiv:1611.05424}, 2016.

\bibitem{noh2015learning}
H.~Noh, S.~Hong, and B.~Han.
\newblock Learning deconvolution network for semantic segmentation.
\newblock In {\em ICCV}, 2015.

\bibitem{oord2016conditional}
A.~v.~d. Oord, N.~Kalchbrenner, O.~Vinyals, L.~Espeholt, A.~Graves, and
  K.~Kavukcuoglu.
\newblock Conditional image generation with {PixelCNN} decoders.
\newblock {\em arXiv:1606.05328}, 2016.

\bibitem{ott2009implicit}
P.~Ott and M.~Everingham.
\newblock Implicit color segmentation features for pedestrian and object
  detection.
\newblock In {\em ICCV}, 2009.

\bibitem{PeronaM90}
P.~Perona and J.~Malik.
\newblock Scale-space and edge detection using anisotropic diffusion.
\newblock {\em PAMI}, 1990.

\bibitem{pohlen2016full}
T.~Pohlen, A.~Hermans, M.~Mathias, and B.~Leibe.
\newblock Full-resolution residual networks for semantic segmentation in street
  scenes.
\newblock In {\em CVPR}, 2014.

\bibitem{spynet}
A.~Ranjan and M.~J. Black.
\newblock Optical flow estimation using a spatial pyramid network.
\newblock {\em CVPR}, 2017.

\bibitem{revaud2015epicflow}
J.~Revaud, P.~Weinzaepfel, Z.~Harchaoui, and C.~Schmid.
\newblock Epic{F}low: {E}dge-preserving interpolation of correspondences for
  optical flow.
\newblock In {\em CVPR}, 2015.

\bibitem{ILSVRC15}
O.~Russakovsky, J.~Deng, H.~Su, J.~Krause, S.~Satheesh, S.~Ma, Z.~Huang,
  A.~Karpathy, A.~Khosla, M.~Bernstein, A.~C. Berg, and L.~Fei-Fei.
\newblock {ImageNet Large Scale Visual Recognition Challenge}.
\newblock {\em IJCV}, 115(3):211--252, 2015.

\bibitem{sermanet-iclr-14}
P.~Sermanet, D.~Eigen, X.~Zhang, M.~Mathieu, R.~Fergus, and Y.~LeCun.
\newblock Over{F}eat: {I}ntegrated recognition, localization and detection
  using convolutional networks.
\newblock In {\em ICLR}, 2014.

\bibitem{shi2000normalized}
J.~Shi and J.~Malik.
\newblock Normalized cuts and image segmentation.
\newblock {\em PAMI}, 22(8):888--905, 2000.

\bibitem{simo2015discriminative}
E.~Simo-Serra, E.~Trulls, L.~Ferraz, I.~Kokkinos, P.~Fua, and F.~Moreno-Noguer.
\newblock Discriminative learning of deep convolutional feature point
  descriptors.
\newblock In {\em ICCV}, 2015.

\bibitem{Simonyan14c}
K.~Simonyan and A.~Zisserman.
\newblock Very deep convolutional networks for large-scale image recognition.
\newblock In {\em ICLR}, 2015.

\bibitem{smith1997susan}
S.~M. Smith and J.~M. Brady.
\newblock {SUSAN} -- {A} new approach to low level image processing.
\newblock {\em IJCV}, 23(1):45--78, 1997.

\bibitem{sukhbaatar2015end}
S.~Sukhbaatar, J.~Weston, R.~Fergus, et~al.
\newblock End-to-end memory networks.
\newblock In {\em NIPS}, pages 2440--2448, 2015.

\bibitem{tomasi1998bilateral}
C.~Tomasi and R.~Manduchi.
\newblock Bilateral filtering for gray and color images.
\newblock In {\em ICCV}, 1998.

\bibitem{trulls2013dense}
E.~Trulls, I.~Kokkinos, A.~Sanfeliu, and F.~Moreno-Noguer.
\newblock Dense segmentation-aware descriptors.
\newblock In {\em CVPR}, 2013.

\bibitem{trulls2014segmentation}
E.~Trulls, S.~Tsogkas, I.~Kokkinos, A.~Sanfeliu, and F.~Moreno-Noguer.
\newblock Segmentation-aware deformable part models.
\newblock In {\em CVPR}, 2014.

\bibitem{weinzaepfel2013deepflow}
P.~Weinzaepfel, J.~Revaud, Z.~Harchaoui, and C.~Schmid.
\newblock Deep{F}low: {L}arge displacement optical flow with deep matching.
\newblock In {\em ICCV}, pages 1385--1392, 2013.

\bibitem{xie15hed}
S.~Xie and Z.~Tu.
\newblock Holistically-nested edge detection.
\newblock In {\em CVPR}, 2015.

\bibitem{fisher}
F.~Yu and V.~Koltun.
\newblock Multi-scale context aggregation by dilated convolutions.
\newblock In {\em ICLR}, 2016.

\bibitem{ZagoruykoCVPR2015}
S.~Zagoruyko and N.~Komodakis.
\newblock Learning to compare image patches via convolutional neural networks.
\newblock In {\em CVPR}, 2015.

\bibitem{vzbontar2014computing}
J.~{\v{Z}}bontar and Y.~LeCun.
\newblock Computing the stereo matching cost with a convolutional neural
  network.
\newblock In {\em CVPR}, 2014.

\bibitem{zheng2015conditional}
S.~Zheng, S.~Jayasumana, B.~Romera-Paredes, V.~Vineet, Z.~Su, D.~Du, C.~Huang,
  and P.~H. Torr.
\newblock Conditional random fields as recurrent neural networks.
\newblock In {\em ICCV}, 2015.

\end{thebibliography}
}

\clearpage
\pagenumbering{arabic} 
\title{Segmentation-Aware Convolutional Networks Using Local Attention Masks\\
Supplementary Material}    

 \twocolumn[{
    \begin{@twocolumnfalse}
      \maketitle
      \includegraphics[width=\linewidth]{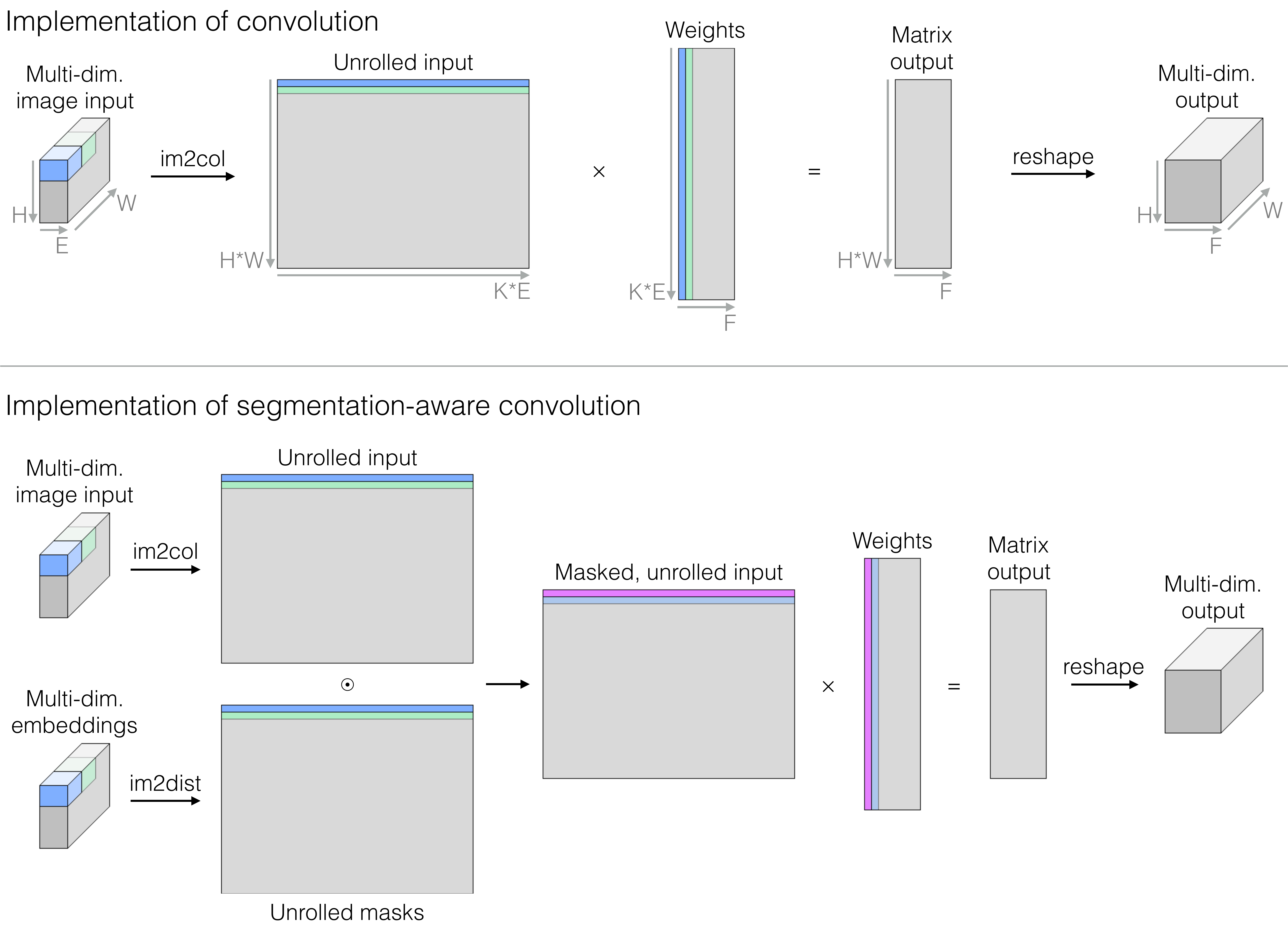}
      \\
      \\
      Figure 1: Implementation of convolution in Caffe, compared with the implementation of segmentation-aware convolution. 
      Convolution involves re-organizing the elements of each (potentially overlapping) patch into a column (\ie, \textit{im2col}), followed by a matrix multiplication with weights. Segmentation-aware convolution works similarly, with an image-to-column transformation on the input, an image-to-distance transformation on the embeddings (\ie, \textit{im2dist}), a pointwise multiplication of those two matrices, and then a matrix multiplication with weights. The variables $H$, $W$ denote the height and width of the input, respectively; $E$ denotes the number of channels in the input; $K$ denotes the dimensionality of a patch (\eg, $K=9$ in convolution with a $3 \times 3$ filter); $F$ denotes the number of filters (and the dimensionality of the output). In both cases, an $H \times W \times E$ input is transformed into an $H \times W \times F$ output. 
    \end{@twocolumnfalse}
 }]

\end{document}